# Phrase-based Machine Translation is State-of-the-Art for Automatic Grammatical Error Correction

**Marcin Junczys-Dowmunt** and **Roman Grundkiewicz**
Adam Mickiewicz University in Poznań
ul. Umultowska 87, 61-614 Poznań, Poland
`{junczys,romang}@amu.edu.pl`

## Abstract

In this work, we study parameter tuning towards the $M^2$ metric, the standard metric for automatic grammar error correction (GEC) tasks. After implementing $M^2$ as a scorer in the Moses tuning framework, we investigate interactions of dense and sparse features, different optimizers, and tuning strategies for the CoNLL-2014 shared task. We notice erratic behavior when optimizing sparse feature weights with $M^2$ and offer partial solutions. We find that a bare-bones phrase-based SMT setup with task-specific parameter-tuning outperforms all previously published results for the CoNLL-2014 test set by a large margin (46.37% $M^2$ over previously 41.75%, by an SMT system with neural features) while being trained on the same, publicly available data. Our newly introduced dense and sparse features widen that gap, and we improve the state-of-the-art to 49.49% $M^2$.

## 1 Introduction

Statistical machine translation (SMT), especially the phrase-based variant, is well established in the field of automatic grammatical error correction (GEC) and systems that are either pure SMT or incorporate SMT as system components occupied top positions in GEC shared tasks for different languages.

With the recent paradigm shift in machine translation towards neural translation models, neural encoder-decoder models are expected to appear in the field of GEC as well, and first published results (Xie et al., 2016) already look promising. As it is the case in classical bilingual machine translation research, these models should be compared against strong SMT baselines. Similarly, system combinations of SMT with classifier-based approaches (Rozovskaya and Roth, 2016) suffer from unnecessarily weak MT base systems which make it hard to assess how large the contribution of the classifier pipelines really is. In this work we provide these baselines.

During our experiments, we find that a bare-bones phrase-based system outperforms the best published results on the CoNLL-2014 test set by a significant margin only due to a task-specific parameter tuning when being trained on the same data as previous systems. When we further investigate the influence of well-known SMT-specific features and introduce new features adapted to the problem of GEC, our final systems outperform the best reported results by 8% $M^2$, moving the state-of-the-art results for the CoNLL-2014 test set from 41.75% $M^2$ to 49.49%.

The paper is organized as follows: section 2 describes previous work, the CoNLL-2014 shared tasks on GEC and follow-up papers. Our main contributions are presented in sections 3 and 4 where we investigate the interaction of parameter tuning towards the $M^2$ metric with task-specific dense and sparse features. Especially tuning for sparse features is more challenging than initially expected, but we describe optimizer hyper-parameters that make sparse feature tuning with $M^2$ feasible. Section 5 reports on the effects of adding a web-scale n-gram language model to our models.

Scripts and models used in this paper are available from https://github.com/grammatical/baselines-emnlp2016 to facilitate reproducibility of our results.

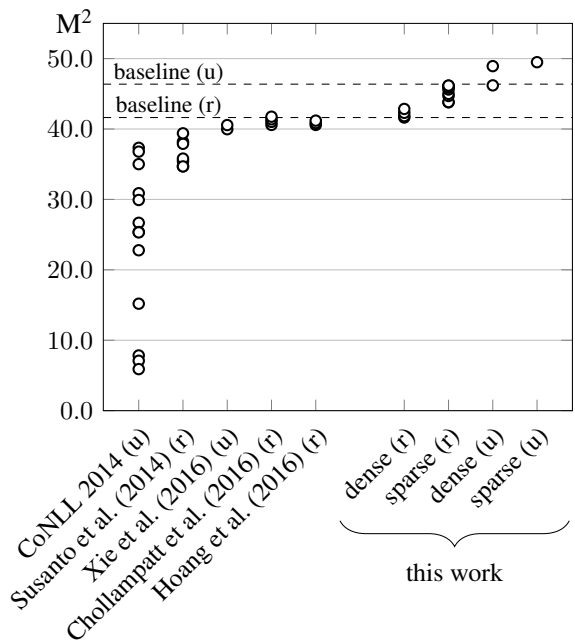

Figure 1: Comparison with previous work on the CoNLL-2014 task, trained on publicly available data. Dashed lines mark results for our baseline systems with restricted (r) and unrestricted (u) data.

## 2 Previous Work

While machine translation has been used for GEC in works as early as Brockett et al. (2006), we start our discussion with the CoNLL-2014 shared task (Ng et al., 2014) where for the first time an unrestricted set of errors had to be fully corrected. Previous work, most notably during the CoNLL shared-task 2013 (Ng et al., 2013), concentrated only on five selected errors types, but machine translation approaches (Yoshimoto et al., 2013; Yuan and Felice, 2013) were used as well.

The goal of the CoNLL-2014 shared task was to evaluate algorithms and systems for automatically correcting grammatical errors in essays written by second language learners of English. Grammatical errors of 28 types were targeted. Participating teams were given training data with manually annotated corrections of grammatical errors and were allowed to use additional publicly available data.

The corrected system outputs were evaluated blindly using the MaxMatch ($M^2$) metric (Dahlmeier and Ng, 2012). Thirteen system submissions took part in the shared task. Among the top-three positioned systems, two submissions — CAMB (Felice et al., 2014) and AMU (Junczys-Dowmunt and Grundkiewicz, 2014)[1] — were partially or fully based on SMT. The second system, CUUI (Rozovskaya et al., 2014), was a classifier-based approach, another popular paradigm in GEC.

After the shared task, Susanto et al. (2014) published work on GEC systems combinations. They combined the output from a classification-based system and a SMT-based system using MEMT (Heafield and Lavie, 2010), reporting new state-of-the-art results for the CoNLL-2014 test set.

Xie et al. (2016) presented a neural network-based approach to GEC. Their method relies on a character-level encoder-decoder recurrent neural network with an attention mechanism. They use data from the public Lang-8 corpus and combine their model with an n-gram language model trained on web-scale Common Crawl data.

More recent results are Chollampatt et al. (2016) and Hoang et al. (2016) which also rely on MT systems with new features (a feed-forward neural translation model) and n-best list re-ranking methods. However, most of the improvement over the CoNLL-2014 shared task of these works stems from using the parameter tuning tools we introduced in Junczys-Dowmunt and Grundkiewicz (2014).

In Figure 1 we give a graphical overview of the published results for the CoNLL-2014 test set in comparison to the results we will discuss in this work. Positions marked with (r) use only restricted data which corresponds to the data set used by Susanto et al. (2014). Positions with (u) make use of web-scale data, this corresponds to the resources used in Xie et al. (2016). We marked the participants of the CoNLL-2014 shared task as unrestricted as some participants made use of Common Crawl data or Google n-grams. The visible plateau for results prior to this work seem to confirm our claims about missing strong baselines.

---

[1] Junczys-Dowmunt and Grundkiewicz (2014) is our own contribution and introduced many of the concepts discussed in this work, but seemingly to little effect during the task. Later analysis revealed that our submission had an incorrectly filtered language model that was missing many possible entries. Our original system without this deficiency would have achieved results around 44% $M^2$ already in 2014. This discovery triggered an intensive reanalysis of our shared task system with significantly new conclusions presented in this work. We apologize for supplying these results so late, as this seems to have halted progress in the field for nearly two years.

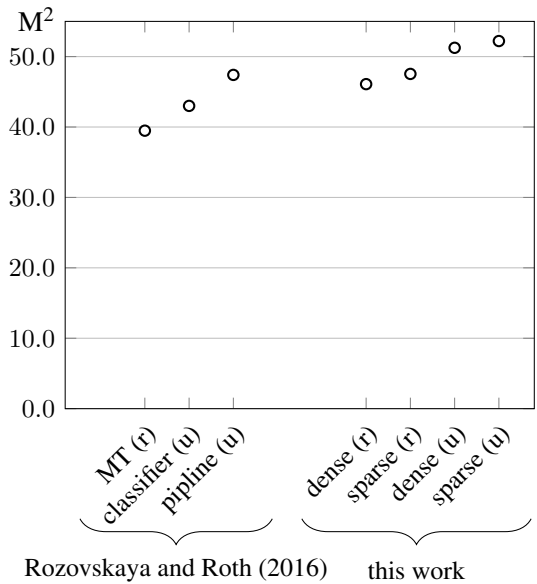

Figure 2: Comparison with Rozovskaya and Roth (2016) using the non-public Lang-8 data set. Here (r) means no web-scale monolingual resources, (u) includes Google 1T n-grams or CommonCrawl.

Rozovskaya and Roth (2016) introduce a SMT-classifier pipeline with state-of-the-art results. Unfortunately, these results are reported for a training set that is not publicly available (data crawled from the Lang-8 website)[2]. Figure 2 compares our results for this resource to Rozovskaya and Roth (2016). See Section 6 for details.

## 3 Dense feature optimization

Moses comes with tools that can tune parameter vectors according to different MT tuning metrics. Prior work used Moses with default settings: minimum error rate training (Och, 2003) towards BLEU (Papineni et al., 2002). BLEU was never designed for grammatical error correction; we find that directly optimizing for $M^2$ works far better.

---

[2]We shared this resource that has been crawled by us for use in Junczys-Dowmunt and Grundkiewicz (2014) privately with Rozovskaya and Roth (2016), but originally were not planning to report results for this resource in the future. We now provide a comparison to Rozovskaya and Roth (2016), but discourage any further use of this unofficial data due to reproducibility issues.

### 3.1 Tuning towards $M^2$

The $M^2$ metric (Dahlmeier and Ng, 2012) is an F-Score, based on the edits extracted from a Levenshtein distance matrix. For the CoNLL-2014 shared task, the $\beta$-parameter was set to 0.5, putting two times more weight on precision than on recall.

In Junczys-Dowmunt and Grundkiewicz (2014) we have shown that tuning with BLEU is counter-productive in a setting where $M^2$ is the evaluation metric. For inherently weak systems this can result in all correction attempts to be disabled, MERT then learns to disallow all changes since they lower the similarity to the reference as determined by BLEU. Systems with better training data, can be tuned with BLEU without suffering this "disabling" effect, but will reach non-optimal performance. However, Susanto et al. (2014) tune the feature weights of their two SMT-based systems with BLEU on the CoNLL-2013 test set and report state-of-the-art results.

Despite tuning with $M^2$, in Junczys-Dowmunt and Grundkiewicz (2014) we were not able to beat systems that did not tune for the task metric. We re-investigated these ideas with radically better results, re-implemented the $M^2$ metric in C++ and added it as a scorer to the Moses parameter optimization framework. Due to this integration we can now tune parameter weights with MERT, PRO or Batch Mira. The inclusion of the latter two enables us to experiment with sparse features.

Based on Clark et al. (2011) concerning the effects of optimizer instability, we report results averaged over five tuning runs. Additionally, we compute parameter weight vector centroids as suggested by Cettolo et al. (2011). They showed that parameter vector centroids averaged over several tuning runs yield similar to or better than average results and reduce variance. We generally confirm this for $M^2$-based tuning.

### 3.2 Dense features

The standard features in SMT have been chosen to help guiding the translation process. In a GEC setting the most natural units seem to be minimal edit operations that can be either counted or modeled in context with varying degrees of generalization. That way, the decoder can be informed on several levels of abstraction how the output differs from the input.[3] In this section we implement several features that try to capture these operation in isolation and in context.

| source phrase | target phrase | LD | D I S |
|---|---|---|---|
| a short time . | short term only . | 3 | 1 1 1 |
| a situation | into a situation | 1 | 0 1 0 |
| a supermarket . | a supermarket . | 0 | 0 0 0 |
| a supermarket . | at a supermarket | 2 | 1 1 0 |
| able | unable | 1 | 0 0 1 |

Table 1: Word-based Levenshtein distance (LD) feature and separated edit operations (D = deletions, I = insertions, S = substitutions)

| Corpus | Sentences | Tokens |
|---|---|---|
| NUCLE | 57.15 K | 1.15 M |
| CoNLL-2013 Test Set | 1.38 K | 29.07 K |
| CoNLL-2014 Test Set | 1.31 K | 30.11 K |
| Lang-8 | 2.23 M | 30.03 M |
| Lang-8 (non-public) | 3.72 M | 51.07 M |
| Wikipedia | 213.08 M | 3.37 G |
| CommonCrawl (u) | 59.13 G | 975.63 G |

Table 2: Parallel (above line) and monolingual training data.

### 3.2.1 Stateless features

Our stateless features are computed during translation option generation before decoding, modeling relations between source and target phrases. They are meant to extend the standard SMT-specific MLE-based phrase and word translation probabilities with meaningful phrase-level information about the correction process.

**Levenshtein distance.** In Junczys-Dowmunt and Grundkiewicz (2014) we use word-based Levenshtein distance between source and target as a translation model feature, Felice et al. (2014) independently experiment with a character-based version.

**Edit operation counts.** We further refine Levenshtein distance feature with edit operation counts. Based on the Levenshtein distance matrix, the numbers of deletions, insertions, and substitutions that transform the source phrase into the target phrase are computed, the sum of these counts is equal to the original Levenshtein distance (see Table 1).

### 3.2.2 Stateful features

Contrary to stateless features, stateful features can look at translation hypotheses outside their own span and take advantage of the constructed target context. The most typical stateful features are language models. In this section, we discuss LM-like features over edit operations.

---

[3]We believe this is important information that currently has not yet been mastered in neural encoder-decoder approaches.

**Operation Sequence Model.** Durrani et al. (2013) introduce Operation Sequence Models in Moses. These models are Markov translation models that in our setting can be interpreted as Markov edition models. Translations between identical words are matches, translations that have different words on source and target sides are substitutions; insertions and deletions are interpreted in the same way as for SMT. Gaps, jumps, and other operations typical for OSMs do not appear as we disabled reordering.

**Word-class language model.** The monolingual Wikipedia data has been used create a 9-gram word-class language model with 200 word-classes produced by word2vec (Mikolov et al., 2013). This features allows to capture possible long distance dependencies and semantical aspects.

### 3.3 Training and Test Data

The training data provided in both shared tasks is the NUS Corpus of Learner English (NUCLE) (Dahlmeier et al., 2013). NUCLE consists of 1,414 essays written by Singaporean students who are non-native speakers of English. The essays cover topics, such as environmental pollution, health care, etc. The grammatical errors in these essays have been hand-corrected by professional English teachers and annotated with one of the 28 predefined error type.

Another 50 essays, collected and annotated similarly as NUCLE, were used in both CoNLL GEC shared tasks as blind test data. The CoNLL-2014 test set has been annotated by two human annotators, the CoNLL-2013 by one annotator. Many participants of CoNLL-2014 shared task used the test set from 2013 as development set for their systems.

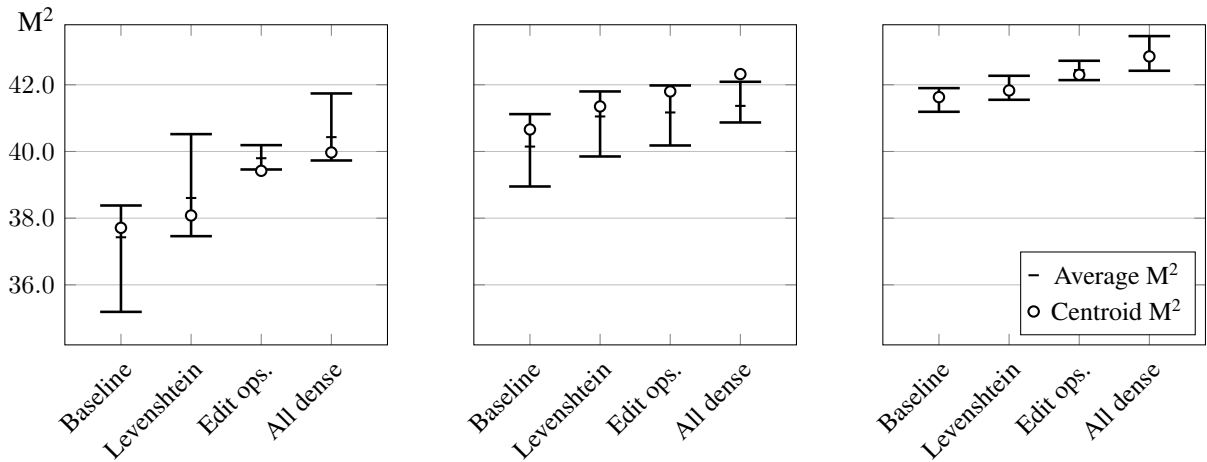

(a) Optimized using BLEU on the CoNLL-2013 test set
(b) Optimized using $M^2$ on the CoNLL-2013 test set
(c) Optimized using $M^2$ on 4 folds of error-rate-adapted NUCLE

Figure 3: Results on the CoNLL-2014 test set for different optimization settings (5 runs for each system) and different feature sets, the "All dense" entry includes OSM, the word class language model, and edit operations). The small circle marks results for averaged weights vectors and is chosen as the final result.

As mentioned before, we report main results using similar training data as Susanto et al. (2014). We refer to this setting that as the "restricted-data setting" (r). Parallel data for translation model training is adapted from the above mentioned NUCLE corpus and the publicly available Lang-8 corpus (Mizumoto et al., 2012), this corpus is distinct from the non-public web-crawled data described in Section 6. Uncorrected sentences serve as source data, corrected counterparts as target data. For language modeling, the target language sentences of both parallel resources are used, additionally we extract all text from the English Wikipedia.

Phrase-based SMT makes it ease to scale up in terms of training data, especially in the case of n-gram language models. To demonstrate the ease of data integration we propose an "unrestricted setting" (u) based on the data used in Junczys-Dowmunt and Grundkiewicz (2014), one of the shared task submissions, and later in Xie et al. (2016). We use Common Crawl data made-available by Buck et al. (2014).

### 3.4 Experiments

Our system is based on the phrase-based part of the statistical machine translation system Moses (Koehn et al., 2007). Only plain text data is used for language model and translation model training. External linguistic knowledge is introduced during parameter tuning as the tuning metric relies on the error annotation present in NUCLE. The translation model is built with the standard Moses training script, word-alignment models are produced with MGIZA++ (Gao and Vogel, 2008), we restrict the word alignment training to 5 iterations of Model 1 and 5 iterations of the HMM-Model. No reordering models are used, the distortion limit is set to 0, effectively prohibiting any reordering. All systems use one 5-gram language model that has been estimated from the target side of the parallel data available for translation model training. Another 5-gram language model trained on Wikipedia in the restricted setting or on Common Crawl in the unrestricted case.

Systems are retuned when new features of any type are added. We first successfully reproduce results from Susanto et al. (2014) for BLEU-based tuning on the CoNLL-2013 test set as the development set (Fig. 3a) using similar training data. Repeated tuning places the scores reported by Susanto et al. (2014) for their SMT-ML combinations (37.90 – 39.39) within the range of possible values for a purely Moses-based system without any specific features (35.19 – 38.38) or with just the Levenshtein distance features (37.46 – 40.52). Since Su-santo et al. (2014) do not report results for multiple tuning steps, the extend of influence of optimizer instability on their experiments remains unclear. Even with BLEU-based tuning, we can see significant improvements when replacing Levenshtein distance with the finer-grained edit operations, and another performance jump with additional stateful features. The value range of the different tuning runs for the last feature set includes the currently best-performing system (Xie et al. (2016) with 40.56%), but the result for the averaged centroid are inferior.

Tuning directly with $M^2$ (Fig. 3b) and averaging weights across five iterations, yields between 40.66% $M^2$ for a vanilla Moses system and 42.32% for a system with all described dense features. Results seen to be more stable. Averaging weight vectors across runs to produce the final vector seems like a fair bet. Performance with the averaged weight vectors is either similar to or better than the average number for five runs.

### 3.5 Larger development sets

No less important than choosing the correct tuning metric is a good choice of the development set. Among MT researches, there is a number of more or less well known truths about suitable development sets for translation-focused settings: usually they consist of between 2000 and 3000 sentences, they should be a good representation of the testing data, sparse features require more sentences or more references, etc. Until now, we followed the seemingly obvious approach from Susanto et al. (2014) to tune on the CoNLL-2013 test set. The CoNLL-2013 test set consists of 1380 sentences, which might be barely enough for a translation-task, and it is unclear how to quantify it in the context of grammar correction. Furthermore, calculating the error rate in this set reveals that only 14.97% of the tokens are part of an erroneous fragment, for the rest, input and reference data are identical. Intuitively, this seems to be very little significant data for tuning an SMT system.

We therefore decide to take advantage of the entire NUCLE data as a development set which so far has only been used as translation model training data. NUCLE consist of more than 57,000 sentences, however, the error rate is significantly lower than in the previous development set, only 6.23%. We adapt the error rate by greedily removing sentences from NUCLE until an error rate of ca. 15% is reached, 23381 sentences and most error annotations remain. We further divide the data into four folds. Each folds serves as development set for parameter tuning, while the three remaining parts are treated as translation model training data. The full Lang-8 data is concatenated with is NUCLE training set, and four models are trained. Tuning is then performed four times and the resulting four parameter weight vectors are averaged into a single weight vector across folds. We repeat this procedure again five times which results in 20 separate tuning steps. Results on the CoNLL-2014 test set are obtained using the full translation model with a parameter vector average across five runs. The CoNLL-2013 test set is not being used for tuning and can serve as a second test set.

As can be seen in Fig. 3c, this procedure significantly improves performance, also for the bare-bones set-up (41.63%). The lower variance between iterations is an effect of averaging across folds.

It turns out that what was meant to be a strong baseline, is actually among the strongest systems reported for this task, outperformed only by the further improvements over this baseline presented in this work.

## 4 Sparse Features

We saw that introducing finer-grained edit operations improved performance. The natural evolution of that idea are features that describe specific correction operations with and without context. This can be accomplished with sparse features, but tuning sparse features according to the $M^2$ metric poses unexpected problems.

### 4.1 Optimizing for $M^2$ with PRO and Mira

The MERT tool included in Moses cannot handle parameter tuning with sparse feature weights and one of the other optimizers available in Moses has to be used. We first experimented with both, PRO (Hopkins and May, 2011) and Batch Mira (Cherry and Foster, 2012), for the dense features only, and found PRO and Batch Mira with standard settings to either severely underperform in comparison to MERT or to suffer from instability with regard to different test sets (Table 3).

| Optimizer | 2013 | 2014 |
|---|---|---|
| MERT | 33.50 | 42.85 |
| PRO | 33.68 | 40.34 |
| Mira | 29.19 | 34.13 |
| -model-bg | 31.06 | 43.88 |
| -D 0.001 | 33.86 | 42.91 |

Table 3: Tuning with different optimizers with dense features only, results are given for the CoNLL-2013 and CoNLL-2014 test set

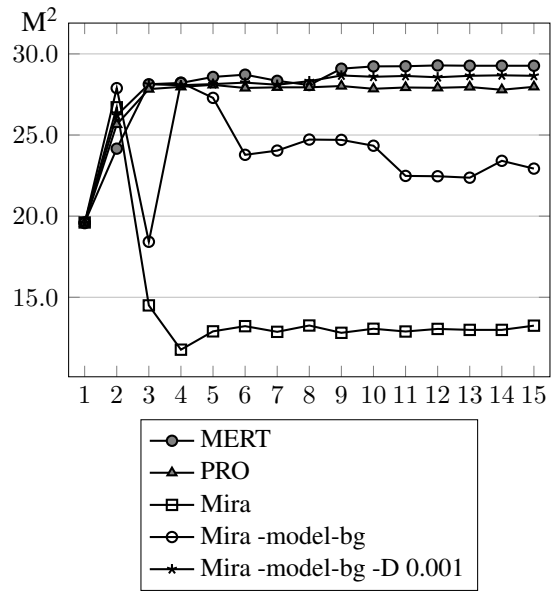

Figure 4: Results per iteration on development set (4-th NUCLE fold)

Experiments with Mira hyper-parameters allowed to counter these effects. We first change the background BLEU approximation method in Batch Mira to use model-best hypotheses (`--model-bg`) which seems to produce more satisfactory results. Inspecting the tuning process, however, reveals problems with this setting, too. Figure 4 documents how instable the tuning process with Mira is across iterations. The best result is reached after only three iterations. In a setting with sparse features this would result in only a small set of weighted sparse features.

After consulting with one of the authors of Batch-Mira, we set the background corpus decay rate to 0.001 (`-D 0.001`), resulting in a sentence-level approximation of $M^2$. Mira's behavior seems to stabilize across iterations. At this point it is not quite clear why this is required. While PRO's behavior is more sane during tuning, results on the test sets are subpar. It seems that no comparable hyper-parameter settings exist for PRO.

### 4.2 Sparse edit operations

Our sparse edit operations are again based on the Levenshtein distance matrix and count specific edits that are annotated with the source and target tokens that took part in the edit. For the following erroneous/corrected sentence pair

```
Err: Then a new problem comes out .
Cor: Hence , a new problem surfaces .
```

we generate sparse features that model contextless edits (matches are omitted):

```
subst(Then,Hence)=1
insert(,)=1
subst(comes, surfaces)=1
del(out)=1
```

and sparse features with one-sided left or right or two-sided context:

```
<s>_subst(Then,Hence)=1
subst(Then,Hence)_a=1
Hence_insert(,)=1
insert(,)_a=1
problem_subst(comes, surfaces)=1
subst(comes, surfaces)_out=1
comes_del(out)=1
del(out)_.=1
<s>_subst(Then,Hence)_a=1
Hence_insert(,)_a=1
problem_subst(comes, surfaces)_out=1
comes_del(out)_.=1
```

All sparse feature types are added on-top of our best dense-features system. When using sparse features with context, the contextless features are included. The context annotation comes from the erroneous source sentence, not from the corrected target sentence. We further investigate different source factors: elements taking part in the edit operation or appearing in the context can either be word forms (factor 0) or word classes (factor 1). As before for dense features we average sparse feature weights across folds and multiple tuning runs.

Figure 5 summarizes the results for our sparse feature experiments. On both test sets we can see significant improvements when including edit-based sparse features, the performance increases even more when source context is added. The CoNLL-2013 test set contains annotations from only one annotator and is strongly biased towards high precision which might explain the greater instability. It appears that sparse features with context where surface forms and word-classes are mixed allow for the best fine-tuning.

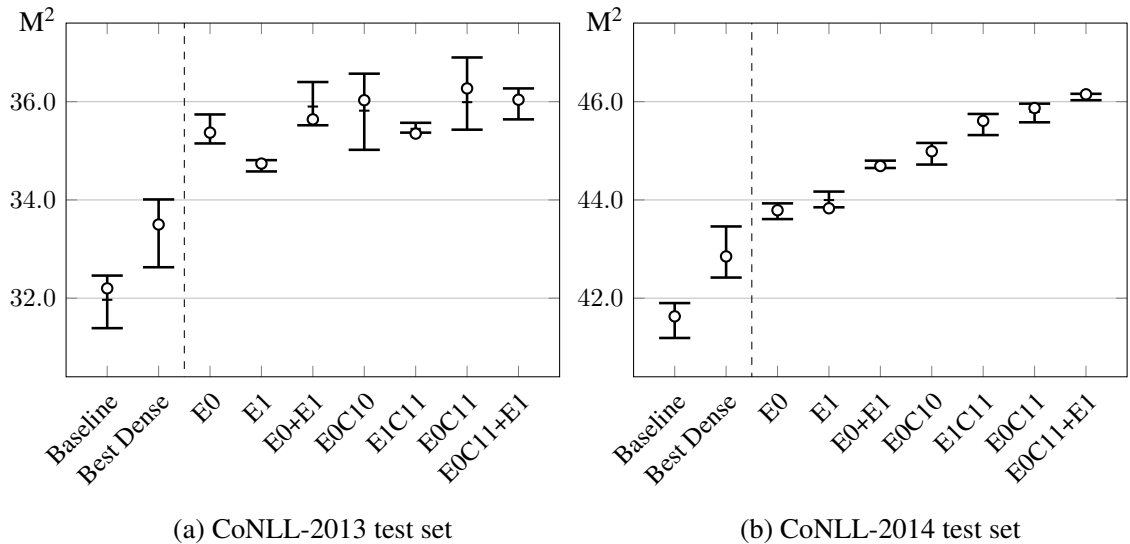

(a) CoNLL-2013 test set      (b) CoNLL-2014 test set

| Symbol | Description |
| --- | --- |
| E0 | Edit operation on words, no context |
| E1 | Edit operation on word classes, no context |
| E0C10 | Edit operation on words with left/right context of maximum length 1 on words |
| E1C11 | Edit operation on word classes with left/right context of maximum length 1 on word classes |
| E0C11 | Edit operation on words with left/right context of maximum length 1 on word classes |

Figure 5: Results on the CoNLL-2014 test set for different sparse features sets

## 5 Adding a web-scale language model

Until now we restricted our experiments to data used by Susanto et al. (2014). However, systems from the CoNLL-2014 were free to use any publicly available data, for instance in Junczys-Dowmunt and Grundkiewicz (2014), we made use of an n-gram language model trained from Common Crawl. Xie et al. (2016) reach the best published result for the task (before this work) by integrating a similar n-gram language model with their neural approach.

We filter the English resources made available by Buck et al. (2014) with cross-entropy filtering (Moore and Lewis, 2010) using the corrected NUCLE corpus as seed data. We keep all sentence with a negative cross-entropy score and compute a 5-gram KenLM (Heafield, 2011) language model with heavy pruning. This step produces roughly 300G of compressed text and a manageable 21G binary model (available for download).

Table 4 summarizes the best results reported in this paper for the CoNLL-2014 test set (column 2014) before and after adding the Common Crawl n-gram language model. The vanilla Moses baseline with the Common Crawl model can be seen as a new simple baseline for unrestricted settings and is ahead of any previously published result. The combination of sparse features and web-scale monolingual data marks our best result, outperforming previously published results by 8% $M^2$ using similar training data. While our sparse features cause a respectable gain when used with the smaller language model, the web-scale language model seems to cancel out part of the effect.

| System | 2014 | | | 2014-10 | | |
| --- | --- | --- | --- | --- | --- | --- |
| | Prec. | Recall | M$^2$ | Prec. | Recall | M$^2$ |
| Baseline | 48.97 | 26.03 | 41.63 | 69.29 | 31.35 | 55.78 |
| +CCLM | 58.91 | 25.05 | 46.37 | 77.17 | 29.38 | 58.23 |
| Best dense | 50.94 | 26.21 | 42.85 | 71.21 | 31.70 | 57.00 |
| +CCLM | 59.98 | 28.17 | 48.93 | 79.98 | 32.76 | 62.08 |
| Best sparse | 57.99 | 25.11 | 45.95 | 76.61 | 29.74 | 58.25 |
| +CCLM | 61.27 | 27.98 | 49.49 | 80.93 | 32.47 | 62.33 |

Table 4: Best results in restricted setting with added unrestricted language model for original (2014) and extended (2014-10) CoNLL test set (trained with public data only).

| System | Prec. | Recall | M$^2$ |
| --- | --- | --- | --- |
| R&R (np) | 60.17 | 25.64 | 47.40 |
| Best dense (np) | 53.56 | 29.59 | 46.09 |
| +CCLM | 61.74 | 30.51 | 51.25 |
| Best sparse (np) | 58.57 | 27.11 | 47.54 |
| +CCLM | 63.52 | 30.49 | 52.21 |

Table 5: Previous best systems trained with non-public (np) error-corrected data for comparison with Rozovskaya and Roth (2016) denoted as R&R.

Bryant and Ng (2015) extended the CoNLL-2014 test set with additional annotations from two to ten annotators. We report results for this valuable resource (column 2014-10) as well.[4] According to Bryant and Ng (2015), human annotators seem to reach on average 72.58% M$^2$ which can be seen as an upper-bound for the task. In this work, we made a large step towards this upper-bound.

## 6 More error-corrected data

As mentioned before, Rozovskaya and Roth (2016) trained their systems on crawled data from the Lang-8 website that has been collect by us for our submission to the CoNLL-2014 shared task. Since this data has not been made officially available, we treat it as non-public. This makes it difficult to put their results in relation with previously published work, but we can at least provide a comparison for our systems. As our strongest MT-only systems trained on public data already outperform the pipelined approaches from Rozovskaya and Roth (2016), it is unsurprising that adding more error-corrected parallel data results in an even wider gap (Table 5). We can assume that this gap would persist if only public data had been used. Although these are the highest reported results for the CoNLL-2014 shared task so far, we think of them as unofficial results and refer to Table 4 as our final results in this work.

## 7 Conclusions

Despite the fact that statistical machine translation approaches are among the most popular methods in automatic grammatical error correction, few papers that report results for the CoNLL-2014 test set seem to have exploited its full potential. An important aspect when training SMT systems that one needs to tune parameters towards the task evaluation metric seems to have been under-explored.

We have shown that a pure SMT system actually outperforms the best reported results for any paradigm in GEC if correct parameter tuning is performed. With this tuning mechanism available, task-specific features have been explored that bring further significant improvements, putting phrase-based SMT ahead of other approaches by a large margin. None of the explored features require complicated pipelines or re-ranking mechanisms. Instead they are a natural part of the log-linear model in phrase-based SMT. It is therefore quite easy to reproduce our results and the presented systems may serve as new baselines for automatic grammatical error correction. Our systems and scripts have been made available for better reproducibility.

---

[4] See Bryant and Ng (2015) for a re-assessment of the CoNLL-2014 systems with this extended test set.

## Acknowledgments

The authors would like to thank Colin Cherry for his help with Batch Mira hyper-parameters and Kenneth Heafield for many helpful comments and discussions. This work was partially funded by the Polish National Science Centre (Grant No. 2014/15/N/ST6/02330) and by Facebook. The views and conclusions contained herein are those of the authors and should not be interpreted as necessarily representing the official policies or endorsements, either expressed or implied, of Facebook.